\documentclass[10pt,twocolumn,letterpaper]{article}

\usepackage{ijcb}
\usepackage{times}
\usepackage{epsfig}
\usepackage{graphicx}
\usepackage{amsmath}
\usepackage{amssymb}

\usepackage{url}

\usepackage{multirow}
\usepackage{booktabs}

\ijcbfinalcopy 

\ifijcbfinal\fi
\begin{document}

\title{Identifying Rhythmic Patterns for Face Forgery Detection and Categorization}

\author{Jiahao Liang\\
	Beijing University of\\Posts and Telecommunications\\
	{\tt\small jiahao.liang@bupt.edu.cn}
	\and
	Weihong Deng\\
	Beijing University of \\Posts and Telecommunications\\
	{\tt\small whdeng@bupt.edu.cn}
}

\maketitle
\thispagestyle{empty}

\begin{abstract}
	With the emergence of GAN, face forgery technologies have been
	heavily abused.
	Achieving accurate face forgery detection is imminent.
	Inspired by remote photoplethysmography (rPPG) that PPG signal
	corresponds to the periodic change of skin color caused by
	heartbeat in face videos, we observe that
	despite the inevitable loss of PPG signal during the forgery process,
	there is still a mixture of PPG signals in the forgery video with a
	unique rhythmic pattern depending on its generation method. Motivated by this
	key observation, we propose a framework for face forgery detection and categorization consisting of:
	1) a Spatial-Temporal Filtering Network (STFNet) for PPG signals filtering,
	and 2) a Spatial-Temporal Interaction Network (STINet) for constraint
	and interaction of PPG signals.
	Moreover,
	with insight into the generation of forgery methods, we further propose
	intra-source and inter-source blending to boost the performance of the
	framework.
	Overall, extensive experiments have proved the superiority of our method.
\end{abstract}

\section{Introduction}
In the past decade, with the rapid development of digital cameras, communication
technologies and mobile devices, video has become a way of entertainment and communication
in people's lives. At the same time,
face forgery technology represented by DeepFakes~\cite{DeepFakes_github} flooded
the network.
Unfortunately, due to its extremely
low barriers and wide accessibility, this technology has been maliciously applied
to create pornographic content~\cite{deepfake_porn}, fake news, financial
fraud and smear politicians. Its unrestricted use
not only damages privacy, law, and politics but also causes a serious crisis of
trust in society.

Therefore, achieving accurate face forgery detection is imminent. Some
studies~\cite{li2018exposing} obtain high accuracy temporarily by learning the pixel differences
between real and forgery images, while face forgery
technology will continue to improve according to the pixel difference as the
supervision information until evading detection, just like GAN generators and
discriminators, competing with each other.
Therefore, to break this deadlock, we need to comprehensively exploit the
differences between real and fake videos, and grasp the essence and core forgery cues
to detect.

Heart rate, as a significant physiological signal in the human body, has a certain
regularity in a period of continuous time. The blood flow caused by the heartbeat forms
periodic changes in the capillaries of human skin tissues, resulting in periodic
changes in the absorbed light and reflected light, and these changes reflecting
the heart rate signals can be
analyzed and acquired in the face video~\cite{verkruysse2008remote}.
The related technology is called remote photoplethysmography (rPPG),
which is widely used in heart rate measurement and face anti-spoofing.

\begin{figure}
\begin{center}
\includegraphics[width=0.95\linewidth]{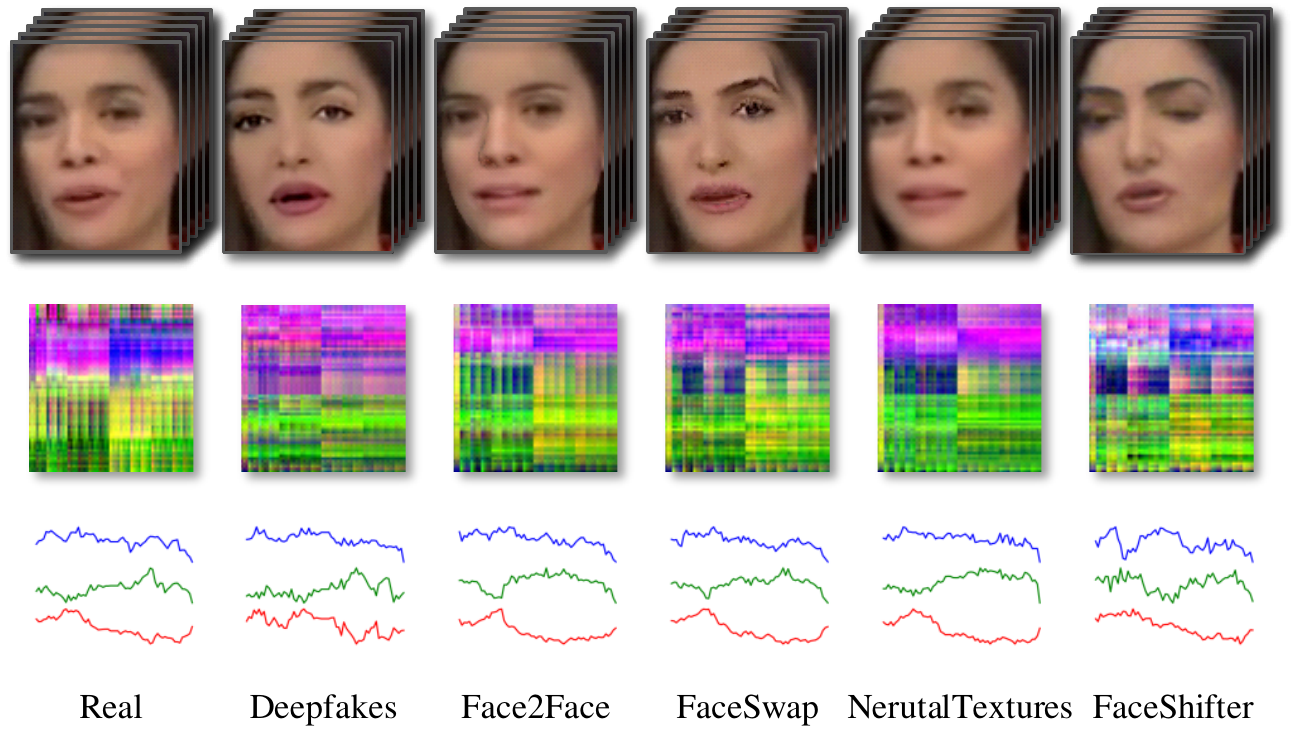}
\end{center}
\caption{An example of spatial-temporal maps (second row) and PPG signals (third row) in the six sub-datasets of FaceForensics++. It can be observed that the presence and unique rhythmic patterns of PPG signals for different face forgery methods.}
\label{fig:source}
\end{figure}


In this paper, we grasp the unique rhythmic pattern of the PPG signal for forgery detection.
Motivated by the key observation that despite the inevitable loss of PPG signal
during the forgery process, there is still a mixture of PPG signals in the forgery
video with a unique rhythmic pattern depending on its generation method,
as shown in Figure~\ref{fig:source}.
Therefore, we design a framework for face forgery detection and categorization.
We use a Spatial-Temporal
Filtering Network to filter out the irrelevant noise in the PPG signal. To
further exploit the potential connections between adjacent PPG signals,
we propose a Spatial-Temporal Interaction Network to interact and
constrain the PPG signals through bidirectional LSTM and adjacency constraint,
respectively.
Moreover,
with insight into the generation of forgery methods, we further propose intra-source
and inter-source blending to boost the performance of the framework.
Overall, extensive experiments have proved the superiority of our method, and our method
achieves remarkable elevation on both face forgery detection and categorization,
but also performs strong robustness and generalization performance.


The contributions of this paper could be summarized as three-fold:
\begin{itemize}
	\vspace{-0.5em}
	\setlength\itemsep{-0.5em}
	\item We propose
		a framework for face forgery detection and categorization. A Spatial-Temporal Filtering
		Network (STFNet) for PPG signals filtering, and a Spatial-Temporal Interaction
		Network (STINet) for interaction and constraint of the PPG signal.
	\item With insight into the generation of forgery methods, we further propose intra-source
		and inter-source data blending to boost the performance of the framework.
	\item Our method
		not only achieves remarkable elevation on both face forgery detection and
		categorization, but also performs strong robustness and
		generalization capabilities.
\end{itemize}

%
%

\section{Related Work}
%
%
\subsection{Face Forgery Generation}

Face forgery has received considerable attention over the past decade.
In the earlier studies, graphics-based methods were used to implement this
technology~\cite{FaceSwap_graphics_url, dale2011video}.
Although graphics-based methods have been evolving, they have not yet become
widespread due to their complexity and huge cost. With the publication of GANs
by Goodfellow \etal~\cite{goodfellow2014generative} in 2014, it has made
impressive progress and attracted a lot of research.
Choi \etal~\cite{choi2018stargan} proposed StarGAN which performs multi-domain
image conversion tasks in one dataset. To address this dilemma that GAN is hard
and slow to train, Karras \etal~\cite{karras2017progressive} proposed ProGAN which
gradually generates images from low quality to high quality.
Overall, These GAN-based methods can easily use some random noise to create
indistinguishable images, which makes face
forgery easier and wildly available to the public.

\subsection{Face Forgery Detection}

Due to the incredible development of deep learning and GAN, face forgery
technology has been progressing rapidly.
To avoid its illegal use, researchers are exploring how to
identify it. In earlier studies, researchers used hand-crafted
features~\cite{pan2012exposing, fridrich2012rich} to detect.
However, these methods don't perform good generalization ability.
In recent years, some methods with automatic learning ability based on deep learning
have been proposed.
Ding \etal~\cite{ding2020swapped} proposed a transfer learning model to
detect swapped faces.
Liu \etal~
\cite{liu2020global} proposed an architecture that can use textures in combination with
the Gram Block module.
Li \etal~\cite{li2020face} assumed that
the face is a blend of two different images, and transformed the task to detect
its stitching boundary for better generalization.
Although these methods achieve a high accuracy rate, they do not take advantage of
deeper features and may fail in the face of new generations of forgery schemes.

\subsection{Remote Photoplethysmography} \label{rPPG}
With the proposal of remote photoplethysmography~\cite{verkruysse2008remote} (rPPG),
a number of heart rate measurement methods based on it have been
developed~\cite{niu2019rhythmnet, yu2019remote, niu2020video}.
It is precisely because of the uniqueness of rPPG signal
that researchers gradually apply it to the field of face
anti-spoofing~\cite{li2016generalized}.

At the same time, face forgery detection methods based on rPPG have also
been proposed. Ciftci \etal~\cite{ciftci2020fakecatcher} analyzed the biological
signals of fake and real video pairs, constructed a generalized classifier, and
convert the signals into PPG maps to train a simple detection network.
Qi \etal~\cite{qi2020deeprhythm} proposed the motion-magnified spatial-temporal
representation and dual-spatial-temporal attention network to adapt to changing faces
and various fake types. However, these methods do not make full use of the
PPG signal.
To further exploit the advantages of the PPG signal,
our method takes into account the noise in the PPG signals as well as the
interaction and constraints of adjacent PPG signal.

\begin{figure*}
\begin{center}
\includegraphics[width=0.95\linewidth]{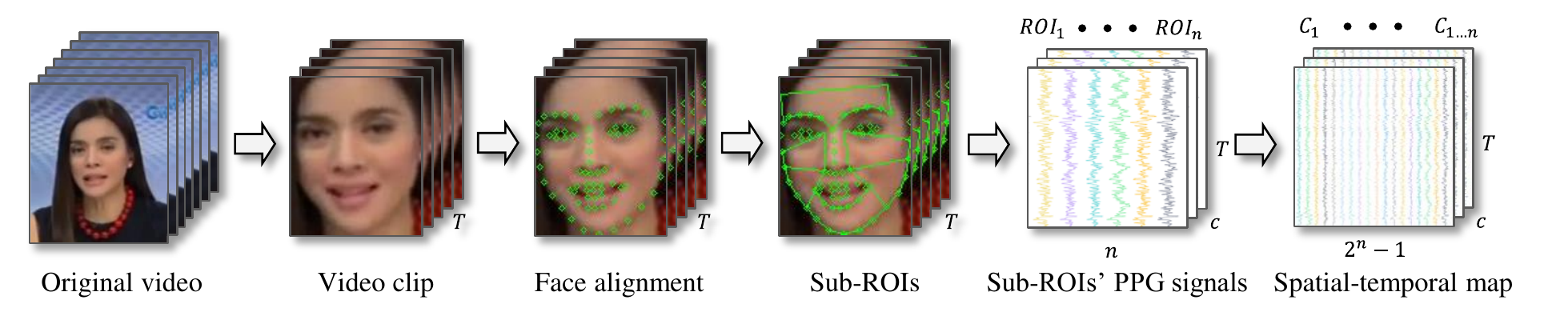}
\end{center}
   \caption{Generation of the spatial-temporal map. For a video,
   we first split out a $T$-frame video clip, next face alignment is implemented
   to locate the landmarks of the face and divide $n$ sub-ROIs according to the
   landmarks, then we calculate the average pixels of each sub-ROI of the $T$-frame video clip.
   Finally, we combine the average
   pixels of $n$ sub-ROIs to obtain a spatial-temporal map with the size of
   $T \times (2^n-1) \times c$. Where $c$ denotes the number of channels, $C_{1 \cdots n}$ denotes the average of the combination of $n$ sub-ROIs' PPG signals.}
\label{fig:spatial-temporal-map}
\end{figure*}

\section{Method}
\subsection{Generation of Spatial Temporal Map}
Since the PPG signal does not simply rely on the texture information of the
images, it cannot be represented directly using the original unprocessed
images. Moreover, the PPG signal is very weak and susceptible to interference.
To enable an accurate and comprehensive representation of the PPG signal,
Inspired by~\cite{niu2020video}, we
generate a spatial-temporal map to represent the PPG signal, as shown in
Figure~\ref{fig:spatial-temporal-map}.

\begin{figure}
\begin{center}
\includegraphics[width=0.85\linewidth]{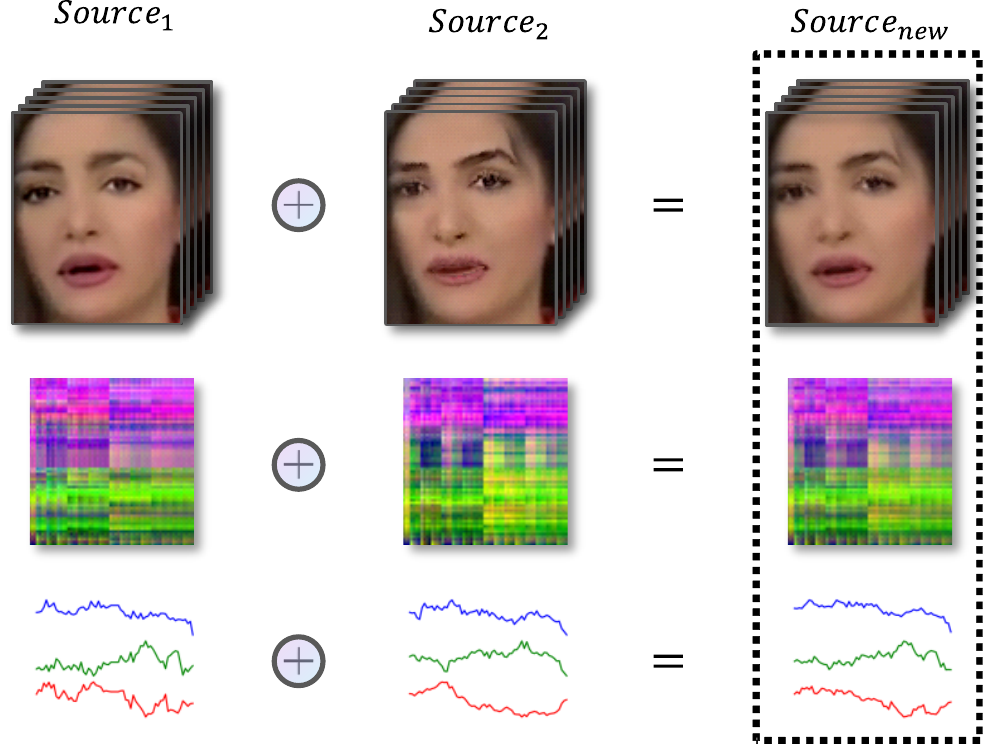}
\end{center}
\caption{An illustration of the data blending. In contrast to the forgery method that process each image (first row) separately, we use the spatial-temporal maps (second row) from two sources to perform blending to generate a new one. The PPG signal (third row) with a unique rhythmic pattern is generated correspondingly.}
   \label{fig:blend}
\end{figure}

\subsection{Intra-Source and Inter-Source Blending}
It is observed that most existing face forgery methods share a common step: different
original face images are transformed and then blended to generate forgery
faces images.
As mentioned in Section~\ref{rPPG}, each face video contains PPG signals corresponding to
the unique heartbeat pattern in the human body. However, the loss of PPG signals is inevitable
during the face forgery generation process. Nonetheless,
these forgery face videos still exist the mixed PPG signals, which are the mixture
of different residual PPG signals that may no longer correspond to human heart rate information
but can reflect the unique rhythmic pattern of the face forgery method. Naturally,
we propose a simple data blending to imitate the face forgery
method for data augmentation.

Unlike forgery methods that process each
image separately,
since the spatial-temporal map is a combination of PPG signals from different
sub-ROIs of the faces in the video clips.
we take into account it and
provide a new approach that is straightforward yet more effective.


Specifically, as shown in Figure~\ref{fig:blend}, given two spatial-temporal maps $M_1$ and $M_2$, we generate
a new spatial-temporal map $M_{new}$ by blending $M_1$ and $M_2$:
\begin{equation}
	M_{new}=M_1 \oplus M_2=\alpha \cdot M_1 + \left(1-\alpha\right) \cdot M_2 \label{blending},
\end{equation}
where $\alpha$ denotes the blending coefficient, which is uniformly distributed between $[0, 1]$.
Moreover, since our detection is a classification task, the data blending can be divided
into \textbf{intra-source} and \textbf{inter-source} blending, \ie, whether the two original spatial-temporal
maps belong to the same source of face forgery or not. Correspondingly, the loss function
of the framework will be different, as detailed in Section~\ref{loss_function}.
Intuitively, intra-source blending expands the data diversity within the source,
while inter-source blending can be considered as expanding the variety of
sources even further.
%
%
\begin{figure*}
\begin{center}
\includegraphics[width=0.95\linewidth]{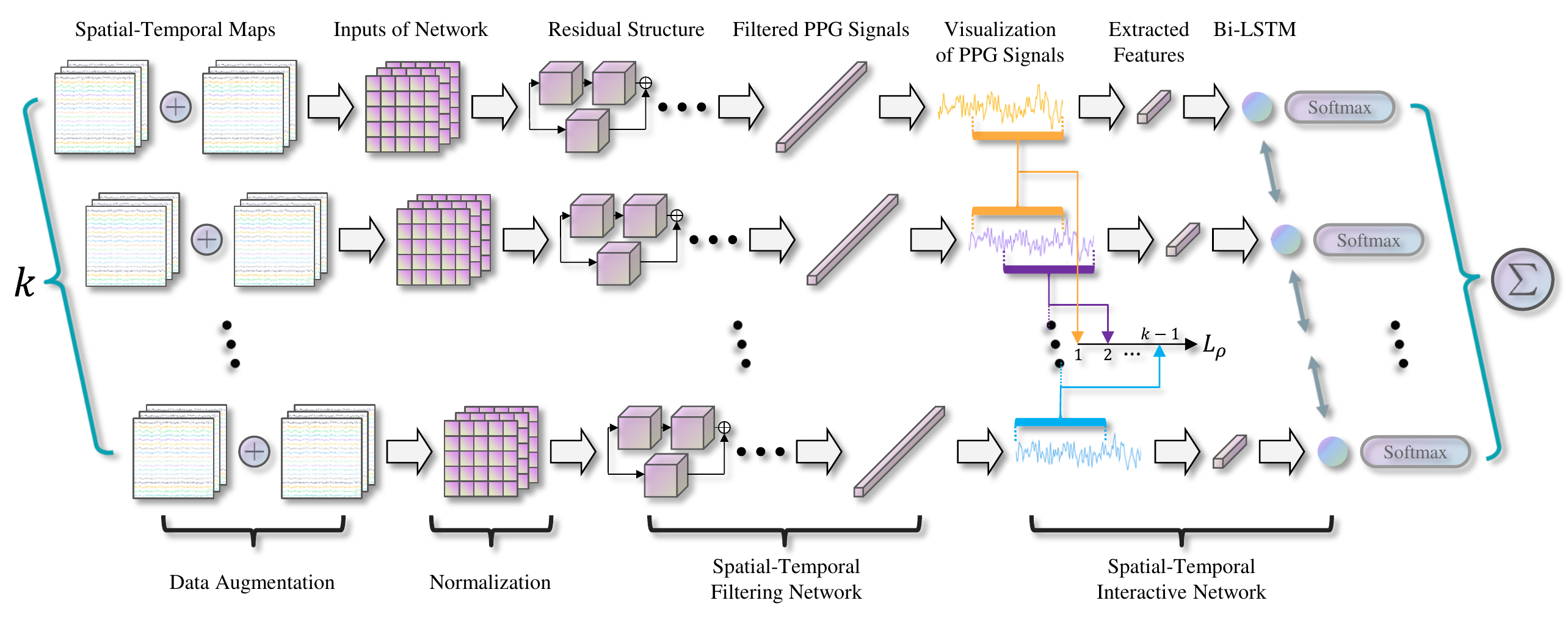}
\end{center}
\caption{The overview framework of our method. We obtain inputs of the network by normalizing the $k$ adjacent spatial-temporal maps after data blending and input them to STFNet to filter PPG signals. Then we constrain and interact with the filtered adjacent PPG signals by adjacency constraint and Bi-LSTM respectively in STINet. Finally, the $k$ outputs are summed to obtain the prediction.}
   \label{fig:network}
\end{figure*}

\subsection{Framework}
In previous research of face forgery detection based on rPPG principle,
the network architectures of some methods are too simple~\cite{ciftci2020fakecatcher} to make
comprehensive use of PPG signal, while some
are too complex~\cite{qi2020deeprhythm}, and require considerable time.
Moreover, since the input of the network is a spatial-temporal map rather than an
ordinary two-dimensional image, the current mainstream network architecture may not
be particularly suitable. Therefore,
we design a framework containing a Spatial-Temporal Filtering Network (STFNet)
for PPG signals filtering and a Spatial-Temporal Interaction Network (STINet)
for constraint and interaction of PPG signals.


\subsubsection{Spatial-Temporal Filtering Network}
Although we can eliminate some noise (\eg, quantization noise of the camera) in the
spatial-temporal map by pooling the ROI, some irrelevant components (\eg, noise caused
by illumination or motion) remain in it.
We need to further filter the correlated interference to improve the signal-to-noise
ratio of the PPG signal.
Some of the previous methods~\cite{de2013robust, de2014improved}
use hand-crafted features based on certain assumptions
in a specific environment to eliminate these irrelevant components. However, these
methods are not robust and may fail in some realistic situations when the assumptions
do not hold. Here without specific assumptions, we propose a
Spatial-Temporal Filtering Network (STFNet) with a residual module to filter out
irrelevant noise in the PPG signal.

The PPG signals are so week that it is
difficult to learn the PPG signal directly from the spatial-temporal map,
while the irrelevant noise signals are relatively more stable.
Inspired by ResNet~\cite{he2016deep}, we design a module with a residual structure
by learning the irrelevant signal in the backbone layer and the difference between
the PPG signal and the irrelevant signal in the shortcut layer, respectively. Finally,
element-wise addition is applied for these two layers to obtain the filtered PPG signal.



\subsubsection{Spatial-Temporal Interaction Network}
The heart rate has a certain periodic pattern, and the PPG signal reflects the heart rate
information. To explore the temporal information in it, we propose
a Spatial-Temporal Interaction Network containing Bi-LSTM and adjacency constraint.

\textbf{Bi-LSTM.}
PPG signals have continuity. Intuitively, we believe that there are some potential
connections between adjacent PPG signals. To explore the temporal information,
we utilize bidirectional LSTM to transfer information from adjacent video clips to
each other. To be specific, we simultaneously input $k$ adjacent spatial-temporal maps into
the STFNet, next we pass the filtered PPG signals
through a convolutional feature extraction network, and then we feed the extracted
features into the one-layer bidirectional LSTM structure. Finally, the mean of the
output of LSTM as our classification probability and the cross-entropy loss is used
as the loss function.
\begin{equation}
	L_{ce}(t)=-\frac{1}{k} \sum_{n=1}^{k}\log \left(\frac{\exp \left(x_t^n\right)}{\Sigma_i \exp \left(x_i^n\right)} \right),
\end{equation}
where $t$ denotes the target face forgery source, $k$ denotes the number of adjacent video clips.


\textbf{Adjacency Constraint.}
Since we use a certain step size $\omega$ to divide the video clips of length $T$
from the video ($\omega < T)$, then the adjacent video clips will have an overlap of
the length of $\widetilde{T} = T - \omega$. Therefore, when $k$ adjacent video clips are
simultaneously fed into the STFNet, the filtered PPG
signals will also have an overlap of length $\widetilde{T}$.
These overlapping parts should have some connection. However, due to preprocessing
such as normalization, these overlapping parts are not necessarily identical
but should still have the similar waveform variations.
To measure the linear similarity error of the overlapping parts,
we use the negative Pearson coefficient to calculate the adjacency loss.
\begin{equation}
	L_{\rho}=1-\frac{1}{k-1} \sum_{i=1}^{k-1} \frac{cov\left(p_i(w \sim T), p_{i+1}(0 \sim \widetilde{T})\right)}{\sigma_{p_i(\omega \sim T)} \sigma_{p_{i+1}(0 \sim \widetilde{T})}},
\end{equation}
where $p_i$ and $p_{i+1}$ denote the PPG signals of two adjacent video clips,
respectively, $cov$ denotes the covariance, $\sigma$ denotes the standard
deviation, $k$ denotes the number of adjacent video clips.

Intuitively, the adjacency loss can be considered as additional supervisory
information, which guides STFNet to filter out the noise of the PPG signal more accurately.
In other words, via adjacency constraint, the STFNet can focus on extracting
PPG signals instead of interference.

\subsection{Loss of the Framework} \label{loss_function}
Overall, we summarize the loss function of our framework.
The final loss function can be written as:
\begin{equation}
	L = \alpha \cdot L_{ce}\left(t_1\right) + \left(1 - \alpha\right) \cdot L_{ce}\left(t_2\right) + \beta \cdot L_{\rho},
\end{equation}
where $t_1$ and $t_2$ denote the two classes of sources, respectively,
$\beta$ is the weight for balancing the loss.
And for intra-source data blending, $t_1$ and $t_2$ belong to the same source and
the loss function can be simplified to:
\begin{equation}
	L = L_{ce}\left(t\right) + \beta \cdot L_{\rho},
\end{equation}
where $t$ denotes the class of source.

\section{Experiments}
\textbf{Datasets.}
To demonstrate the effectiveness of our method, we selected the most challenging
datasets for validation, FaceForensics++~\cite{rossler2019faceforensics++}
and Celeb-DF~\cite{li2020celeb}, from which we evaluated our method.

FaceForensics++ contains 1,000 original videos
downloaded from the internet.
Five different types of forgery video sub-datasets are generated by computer graphics-based
and learning-based methods. \ie, Deepfakes~\cite{deepfakes_faceswap},
Face2Face~\cite{thies2016face2face}, FaceSwap~\cite{FaceSwap_graphics_url},
NeuralTextures ~\cite{thies2019deferred} and FaceShifter. Each sub-dataset contains 1,000 videos.
FaceForensics++ dataset provides three video quality versions, \ie, RAW (raw video), HQ
(constant rate quantization parameter equal to 23), and LQ (constant rate quantization
parameter equal to 40). And it still contains a dataset DFD dedicated to Deepfakes,
containing 363 real videos and 3068 forgery videos.

Celeb-DF used 59 celebrity interview videos on
YouTube as the original videos. In total, 590 real videos and 5,639
DeepFake videos are included.


\textbf{Implementation Details.}
Unless otherwise specified, the dataset is divided into a ratio of 7:3, the number of
frames $T$ of the video segment is $64$, the
number of sub-ROIs $n$ is $6$, the size of the spatial-temporal map is $64 \times 63$,
and the loss balance weight $\beta = 0.1$.
For a complete video, we use a frame step of 16 to get all the video clips.
All the code is based on the pytorch\footnote{https://pytorch.org/} framework and
trained with NVIDIA GTX 1080Ti.
We use SGD as our optimizer with an initial learning rate of 0.1 and a half decay every
10 epochs. The maximum epoch number is 100.

\textbf{Prediction Aggregation.}
Since a video contains multiple video clips, we take the sum of the predicted
probabilities of all video clips as the final predicted probability of the whole video.

\subsection{Results}
\begin{figure}
\begin{center}
	\includegraphics[width=1\linewidth]{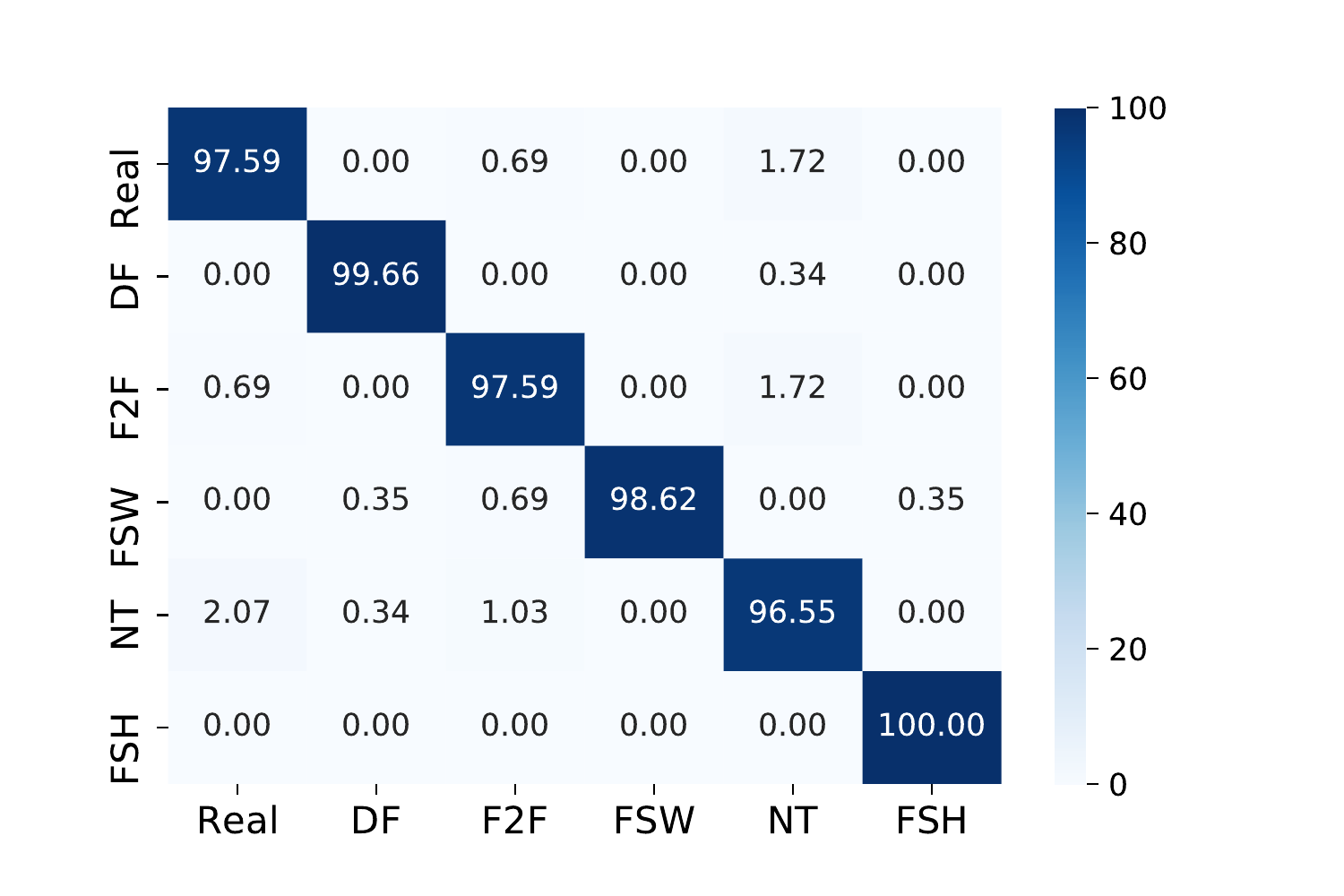}
\end{center}
   \caption{Confusion matrix for face forgery categorization with the average accuracy of 98.33\%.}
\label{fig:confusion_matrix}
\end{figure}

To verify that the presence and unique rhythmic pattern of the PPG signal in each
forgery method, in other words, to verify that the PPG signal of each
face forgery source does have its unique rhythmic pattern, rather than the
PPG signals of all
face forgery sources sharing the same pattern, we conduct pairwise detection
experiments (real-fake and fake-fake) and face forgery categorization experiment
(1 real and 5 fakes) in the six sub-datasets of FaceForensics++, as shown in
Table~\ref{tab:pairwise} and Figure~\ref{fig:confusion_matrix}, respectively.
The results show that we achieve high accuracy in both pairwise detection and face
forgery categorization.

\begin{table*}
	\centering
	\setlength\tabcolsep{7pt}
	\caption{Comparison experiments with other methods in the six sub-datasets of FaceForensics++. The left half is the face forgery binary classification experiment (real-fake), and the right half is the face categorization experiment (1 real and 5 fakes). The metric is accuracy (\%) and the best results are highlighted.}
	\scalebox{0.9}
	{
	\begin{tabular}{lcccccccccccc}
		\toprule
		\multirow{2}{*}{\vspace{-2mm}Method} & \multicolumn{5}{c}{Face forgery detection (real-fake)} &  \multicolumn{7}{c}{Face forgery categorization (1 real and 5 fakes)} \\
		\cmidrule(lr){2-6} \cmidrule(lr){7-13}
		 & DF & F2F & FSW & NT & FSH & Real & DF & F2F & FSW & NT & FSH & Avg \\
		\midrule
		VGG~\cite{simonyan2014very}   & 50.00    & 50.00    & 50.13 & 50.00    & 61.06 & 26.35 & 0.00     & 77.70  & 0.00     & 0.00     & 0.00     & 18.64 \\
		ResNet~\cite{he2016deep} & 89.95 & 77.39 & 74.81 & 74.12 & 82.16 & 97.31 & 98.69 & 94.24 & 92.56 & 88.55 & 98.86 & 95.29 \\
		Inception V3~\cite{szegedy2016rethinking} & 99.74 & 64.36 & 87.74 & 59.8  & 99.35 & 97.76 & 98.32 & 95.49 & 94.22 & 82.8  & 97.96 & 94.77 \\
		Xception~\cite{chollet2017xception} & 99.75 & 98.53 & 96.14 & 91.46 & 99.97 & \textbf{99.84} & 99.59 & \textbf{98.55} & 92.76 & 89.62 & \textbf{100.00}   & 97.11 \\
		\midrule
		Ciftci~\etal~\cite{ciftci2020hearts}   & - & - & - & - & - & 97.29  & 94.66 & 91.66 & 92.33 & 81.93 & - & 93.39 \\
		FakeCatcher~\cite{ciftci2020fakecatcher} & 94.87 & 96.37 & 95.75 & 89.12 & - & - & - & - & - & - & - & - \\
		DeepRhythm~\cite{qi2020deeprhythm} & \textbf{100.00} & \textbf{99.50} & \textbf{100.00} & - & - & - & - & - & - & - & - & - \\
		\midrule
		Our method & \textbf{100.00}   & \textbf{99.50}  & \textbf{100.00}   & \textbf{97.00}  & \textbf{100.00}   & 97.59 & \textbf{99.66} & 97.59 & \textbf{98.62} & \textbf{96.55} & \textbf{100.00}   & \textbf{98.33} \\
		\bottomrule
	\end{tabular}%
	}
	\label{tab:comparision}%
\end{table*}%

\begin{table}
	\centering
	\caption{Results of pairwise detection.}
	\setlength\tabcolsep{6pt}
	\scalebox{0.9}
	{
	\begin{tabular}{lcccccc}
	\toprule
	Dataset & Real & DF & F2F & FSW & NT & FSH \\
	\midrule
		Real  & - & 100.00 & 97.93 & 100.00 & 96.38 & 99.14 \\
		DF & - & - & 99.83 & 99.83 & 99.65 & 99.83 \\
		F2F & - & - & - & 99.48 & 98.28 & 99.66 \\
		FSW & - & - & - & - & 100.00 & 98.48 \\
		NT & - & - & - & - & - & 99.83 \\
		FSH & - & - & - & - & - & - \\
	\bottomrule
	\end{tabular}
	}
	\label{tab:pairwise}
\end{table}

\subsection{Comparison}
To demonstrate the superiority
of our proposed method, we compare it with other methods.
We select the mainstream network, \ie, VGG~\cite{simonyan2014very}, ResNet~\cite{he2016deep},
and some methods that have achieved good performance in FaceForensics++'s benchmark,
\ie, Inception\_V3~ \cite{szegedy2016rethinking}, Xception~\cite{chollet2017xception},
and methods that are also based on the rPPG principle,
\ie, Ciftci\etal~\cite{ciftci2020hearts}, FakeCatcher~\cite{ciftci2020fakecatcher}, DeepRhythm~\cite{qi2020deeprhythm}.

It is worth noting that, except for the method based on the rPPG principle,
the rest of the methods
detect forgery videos by single images rather than video clips, so we adapt the training
strategy for these methods to better compare. Specifically, we directly feed the face region detected by
Dlib with appropriate scaling into the network. To speed up the training, we utilize every
$64^{th}$ frame to train. This can be considered as a baseline for frame-based detection.
\textit{For a fair and comprehensive comparison, in the face forgery detection and face
forgery categorization experiments, we follow DeepRhythm and Ciftci~\etal~\cite{ciftci2020hearts} to divide
the dataset in the ratio of \textbf{8:1:1} and \textbf{7:3}, respectively.}
All methods are experimented in six sub-datasets of FaceForensics++.

\begin{figure*}
\begin{center}
	\includegraphics[width=1\linewidth]{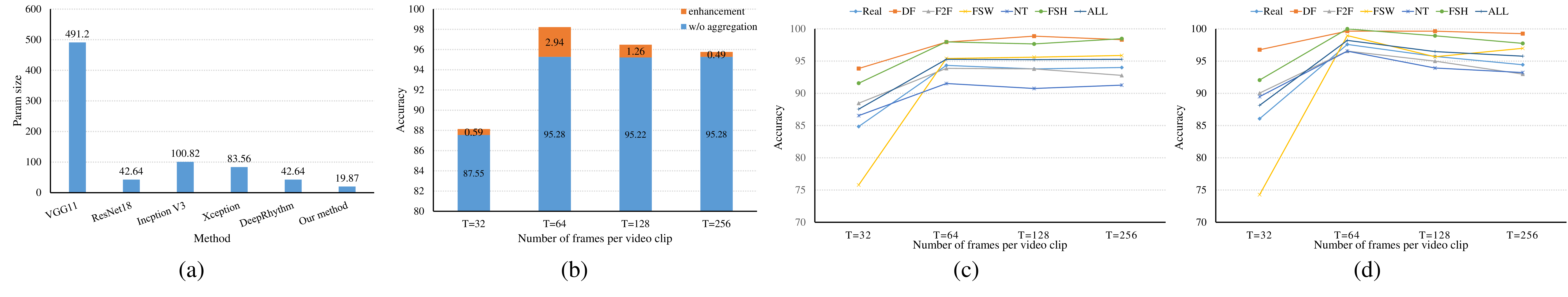}
\end{center}
\caption{(a) Comparison of model parameter size. (b) Accuracy with the single video clip and prediction aggregation. (c) Accuracy of face forgery categorization with single video clip. (d) Accuracy of face forgery categorization with prediction aggregation.}
\label{fig:all}
\end{figure*}

Overall, in Table~\ref{tab:comparision}, we can see that our method
outperforms state-of-the-art methods in both face forgery detection
and categorization, which fully demonstrates the effectiveness of our method.
Baseline methods that do not utilize PPG signals can achieve high accuracy in some datasets.
However, when facing NerualTextures-based rendering videos, these methods that only consider
single-frame spatial information may fail and the accuracy will drop sharply.
This proves that PPG signals have deeper information compared to 2D images, and PPG signals with
spatial-temporal information can represent the more unique difference between real and fake videos.
Compared with methods also based on the rPPG principle,
our method achieves the same state-of-the-art accuracy as DeepRhythm in face forgery detection
experiments (DeepRhythm only tests three sub-datasets).
However,
the model size of our method is much smaller than DeepRhythm (see
Figure~\ref{fig:all} (a)). Moreover, it can be seen that our method gets
significant improvement in the accuracy of face forgery categorization compared
with Ciftci~\etal~\cite{ciftci2020hearts}.
These comparisons demonstrate that our method performs a stronger ability
to exploit the PPG signal.

\subsection{Ablation Experiments}
To demonstrate the effectiveness of our method, based on face forgery categorization
experiments, we perform several ablations to better understand the contributions of each component
in face forgery categorization experiments, as shown in Table~\ref{tab:ablation}.

\begin{table}
	\centering
	\caption{Ablation study of our method by progressively adding the PPG signals (PPG), Spatial-Temporal Filtering Network (STFNet), intra-source data blending (Intra), inter-source data blending, bidirectional LSTM (LSTM), and adjacency constraint (Constraint).}
	\setlength\tabcolsep{4pt}
	\scalebox{0.9}
	{
	\begin{tabular}{ccccccc}
		\toprule
		PPG & STFNet & Intra & Inter & LSTM & Constraint & Avg Acc \\
		\hline
		& \checkmark & & & & & 94.25 \\
		\checkmark & & & & & & 96.04 \\
		\checkmark & \checkmark & & & & & 97.19 \\
		\checkmark & \checkmark & \checkmark & & & & 97.76 \\
		\checkmark & \checkmark & & \checkmark & & & 98.22 \\
		\checkmark & \checkmark & & \checkmark & \checkmark & & 98.27 \\
		\checkmark & \checkmark & & \checkmark & & \checkmark & 98.10 \\
		\checkmark & \checkmark & & \checkmark & \checkmark & \checkmark & \textbf{98.33} \\
		\bottomrule
	\end{tabular}%
	}
	\label{tab:ablation}%
\end{table}%

\textbf{Effectiveness of PPG Signal.}
We feed face images and spatial-temporal maps containing PPG signals into the STFNet, respectively.
the average detection accuracy of utilizing PPG signals is
1.79\% higher than not. It is worth emphasizing that these results demonstrate that PPG
signals from different face forgery sources have their unique rhythmic patterns, and the network can
achieve high detection accuracy by learning their unique rhythmic patterns.

\textbf{Effectiveness of STFNet.}
We utilize STFNet and VGG without classification layers as the backbone
lay of the network for experiments, respectively.
The results show that STFNet outperforms VGG in all aspects and improves the average
accuracy by 1.15\%, proving the effectiveness of the STFNet.

\textbf{Effectiveness of Data Augmentation.}
We further evaluate the effectiveness of our intra-source and inter-source data blending.
The results show that both data blending improve the accuracy. Moreover, inter-source
blending is more effective than intra-source blending, with an improvement of 0.46\%, and
an improvement of 1.03\% compared to no blending, which proves that our data augmentation
can achieve amazing improvement by cleverly blends spatial-temporal maps without any
additional computation and processing.

\textbf{Effectiveness of STINet.}
Finally, we verify the effectiveness of STINet. We first add adjacency constraint and
Bi-LSTM separately, with 0.12\% reduction and 0.05\% improvement in accuracy,
respectively, while the accuracy increases by 0.11\% after combining these two.
After analysis, we speculate that although adding adjacency constraint alone has a certain
constraint effect on the network, it does not play an
optimization role because there is no information interaction between adjacent video clips.
Therefore, combining the two can play a mutually reinforcing role.

\subsection{Extended Experiments}
To verify the generalization performance of our method, we conduct face
forgery categorization experiments by adding the forgery sets of DFD, Celeb, and both of them,
respectively. To save compute, only 300 videos are selected for each dataset
including the six sub-datasets of FaceForensics++.
The results are shown in Table~\ref{tab:extend}, whether the two
datasets are added separately or simultaneously, high accuracy is achieved, which
proves that our method is also effective when other face forgery sources are
expanded.

\begin{table}
	\centering
	\caption{Results of Extended experiment. ``-" represents that the dataset is not added.}
	\setlength\tabcolsep{3pt}
	\scalebox{0.88}
	{
		\begin{tabular}{ccccccccc}
		\toprule
			Real & DF & F2F & FSW & NT & FSH & DFD & Celeb & Avg \\
		\midrule
			96.59 & 98.86 & 94.32 & 97.73 & 98.86 & 97.73 & 98.73 & - & 97.53 \\
			95.45 & 100.00 & 95.45 & 98.86 & 96.59 & 97.73 & - & 98.89 & 97.57 \\
			96.59 & 100.00 & 97.73 & 98.86 & 97.73 & 98.86 & 98.73 & 98.89 & 98.42 \\
		\bottomrule
		\end{tabular}%
	}
	\label{tab:extend}%
\end{table}%

\begin{table}
	\centering
	\caption{Accuracy of face forgery categorization in the compressed datasets of HQ and LQ versions.}
	\setlength\tabcolsep{4pt}
	\scalebox{0.9}
	{
	\begin{tabular}{lccccccc}
		\toprule
		Version  & Real & DF & F2F & FSW & NT & FSH & Avg \\
		\midrule
		HQ    & 92.23 & 99.65 & 93.99 & 85.11 & 85.51 & 97.53 & 92.34 \\
		LQ    & 61.15 & 95.58 & 77.36 & 29.59 & 25.51 & 42.86 & 55.37 \\
		\bottomrule
	\end{tabular}%
	}
	\label{tab:video_compression}%
\end{table}%

\subsection{Video Clip Length and Prediction Aggregation}
To test the performance and improvement of prediction aggregation with different
lengths of video clips. We choose $T = \{32,64,128,256\}$ frames to conduct experiments.

The results are shown in Figure~\ref{fig:all} (b), (c), and (d). We can see that the accuracy of detection of
video clips with $T = \{64,128,256\}$ is very close, 95.28\%, 95.22\%, and 95.28\%,
respectively, while it decreases sharply at $T=\{32\}$, only 87.55\%, which
corresponds to the theoretical fact of rPPG principle: the length of video clips
too short may not contain the complete heart rate cycle and thus fail to extract PPG
signal accurately, while too long may include additional noise. This once again
demonstrates that our method has an extremely strong theoretical basis. Moreover,
prediction aggregation has a certain boost on the accuracy to different
degrees.

\subsection{Video Compression}
To demonstrate the robustness of our proposed method for compressed videos, we conduct experiments
on the HQ version (a light compression) and LQ version(a heavy compression) of FaceForensics++
dataset. As shown in Table~\ref{tab:video_compression}, our method still achieves the average
accuracy of 92.34\% for face forgery categorization
in HQ version, which proves the robustness of our method even in light compression video.
In contrast, in LQ version, due to severe compression, the PPG
signal usually suffers from noisy curve shapes and inaccurate peak locations due to information
loss caused by intra-frame and inter-frame coding of the video compression
process~\cite{yu2019remote}, which leads
our method to obtain only an average accuracy of 55.37\%.

\section{Conclusion}
In this paper,
we propose a framework for face forgery detection and categorization.
It is intuitively motivated by the key observation that despite the inevitable
loss of PPG signal during the forgery process, there is still a mixture of PPG
signals in the forgery video with a unique rhythmic pattern depending on its
generation method. Moreover, with insight into the generation of forgery methods,
we further propose intra-class and inter-class blending to boost the performance
of the framework. Extensive experiments demonstrate the superiority of our method.
Meanwhile, we also explore the impact of video clip length, predictive aggregation,
and video compression.

{\small
	\bibliographystyle{ieee}
	\bibliography{egbib}
}

\end{document}